\documentclass[letterpaper, 10 pt, conference]{ieeeconf}  

\IEEEoverridecommandlockouts                              

\usepackage[utf8]{inputenc} 
\usepackage[T1]{fontenc}    
\usepackage{hyperref}       
\usepackage{url}            
\usepackage{booktabs}       
\usepackage{amsfonts}       
\usepackage{nicefrac}       
\usepackage{microtype}      
\usepackage{graphicx}
\graphicspath{{images/}{../images/}}
\usepackage{amssymb}
\usepackage{amsmath}
\usepackage{xcolor}
\usepackage{bbm}
\usepackage{multirow}
\usepackage{wrapfig}
\usepackage{mathtools}

\DeclareMathOperator*{\argmin}{arg\,min}

\usepackage{xspace}
\makeatletter
\DeclareRobustCommand\onedot{\futurelet\@let@token\@onedot}
\def\@onedot{\ifx\@let@token.\else.\null\fi\xspace}

\def\eg{\emph{e.g}\onedot} 
\def\ie{\emph{i.e}\onedot}

\def\wrt{w.r.t\onedot} 
\def\etal{\emph{et al}\onedot}

\title{\LARGE \bf
Fast Few-Shot Classification by Few-Iteration Meta-Learning
}

\author{Ardhendu Shekhar Tripathi, Martin Danelljan, Luc Van Gool and Radu Timofte
\thanks{All authors are affiliated with the Computer Vision Lab, Department of Electrical Engineering, ETH Zurich, Switzerland.
{\tt\small \{ardhendu-shekhar.tripathi, martin.danelljan, vangool, radu.timofte\}@vision.ee.ethz.ch}}
\thanks{This work was supported by the ETH Zurich Fund (OK), a Huawei Technologies Oy (Finland) project, an Amazon AWS grant, and Nvidia.}
}
\begin{document}
\maketitle
\begin{abstract}
  Autonomous agents interacting with the real world need to learn new concepts efficiently and reliably. This requires learning in a low-data regime, which is a highly challenging problem. We address this task by introducing a fast optimization-based meta-learning method for few-shot classification. It consists of an embedding network, providing a general representation of the image, and a base learner module. The latter learns a linear classifier during the inference through an unrolled optimization procedure. We design an inner learning objective composed of (i) a robust  classification loss on the support set and (ii) an entropy loss, allowing transductive learning from unlabeled query samples. By employing an efficient initialization module and a Steepest Descent based optimization algorithm, our base learner predicts a powerful classifier within only a few iterations. Further, our strategy enables important aspects of the base learner objective to be learned during meta-training. To the best of our knowledge, this work is the first to integrate both induction and transduction into the base learner in an optimization-based meta-learning framework. We perform a comprehensive experimental analysis, demonstrating the speed and effectiveness of our approach on four few-shot classification datasets. The Code is available at \href{https://github.com/4rdhendu/FIML}{\textcolor{blue}{https://github.com/4rdhendu/FIML}}.

\end{abstract}

\section{Introduction}
Unlike humans, most machine learning techniques require thousands of examples to even achieve acceptable performance on a single task. However, we ultimately want autonomous agents that can learn new concepts on-the-fly, given only a few training samples.
This requires efficient online learning from scarce data. The problem of few-shot learning, \ie learning tasks from scarce data, has therefore gained significant interest in recent years~\cite{vinyals2016matching, snell2017prototypical, finn2017maml}.
Among current directions, few-shot classification~\cite{closerlook} aims at learning a classifier given only a few labeled examples.
In the extreme, there might only be a single example of each class.
One promising direction in few-shot learning is to design methods that gain experience from learning to solve other similar tasks in order to better learn the task at hand. 
This is generally referred to as meta-learning~\cite{lake2015human}, aiming to `learning to learn' how to solve individual tasks.


Meta-learning itself covers a wide diversity of methods. In few-shot classification, two types of approaches have been particularly successful, namely metric learning~\cite{vinyals2016matching, snell2017prototypical, sung2018learning} and optimization-based~\cite{finn2017maml,ravi2016optimization,  ImplicitGrad, metaoptnet, luca}. While the former learn an embedding space, the latter aim at optimizing a set of model parameters to solve the specific task.
We build on the optimization-based paradigm, as it allows for the integration of powerful task-specific learning formulations that require objective minimization.
We consider the setting with two main modules, (i) a meta-network and (ii) a base learner. The meta-network generally learns a feature representation across a distribution of tasks. The base learner, on the other hand, performs the task-specific adaptation. To ensure practical meta-learning, the adaptation process must be both, efficient and differentiable. Recent works have explored closed form solutions~\cite{luca} and implicit differentiation of the optimality conditions~\cite{metaoptnet}. However, these methods lack significant flexibility in terms of the choice of base learner objective. In this work, we therefore explore an alternative approach.

 


We propose a base learner employing unrolled optimization. To this end, we utilize the steepest descent algorithm, combined with quadratic approximation and effective initialization, which ensures reliable convergence within only a few iterations. Our optimization strategy allows for greater flexibility in the choice of the base learner objective. Specifically, we formulate a parametrized objective that is \emph{learned} during meta-training. Moreover, as made possible by our meta-learning framework, we investigate the integration of transductive learning strategies into the base learner itself. Unlike previous optimization-based methods, which only employ support images, our base learner incorporate information from the query samples, both, during meta-training and inference. Our meta-learning approach is efficient, achieving 200$\times$ faster inference time compared to state-of-the-art fine-tuning based methods~\cite{dhillon2019baseline}.



\textbf{Contributions: }
Our work contains the following main contributions. (i) We propose FIML, a flexible meta-learning framework based on an efficient unrolled optimization strategy. (ii) We introduce a learnable inductive objective, employed by the base learner. (iii) We integrate transductive learning into the base learner. 
(iv) We develop a method for leveraging spatial dense features for few-shot classification. To validate each of our design choices, we perform extensive ablative experiments. Further, we evaluate our approach on four few-shot classification benchmarks, setting a new state-of-the-art in several settings.



\section{Related Work}
Current state-of-the-art meta-learning methods can be broadly categorized into two main categories - metric-based and optimization-based. The metric-based methods~\cite{vinyals2016matching, snell2017prototypical, sung2018learning, koch2015siamese} aim at learning a common embedding space where it is easy to distinguish between different categories through a distance metric. Optimization-based meta-learning methods aim to learn to adapt a model to a given task. Notably, a family of MAML~\cite{finn2017maml} based methods adapt a set of prior model parameters to the task through first-order optimization ~\cite{first_order_reptile}. Recently, Zintgraf \etal ~\cite{zintgraf2018cavia} proposed an extension to MAML called CAVIA that is less prone to meta-overfitting. CAVIA partitions the model parameter into task-specific context parameters and the task-agnostic shared parameters and updates only the context parameters at test time. In a similar direction, Rajeswaran \etal~\cite{ImplicitGrad} suggest an implicit MAML variant that mitigates the issue  of differentiating through the task-specific inner-learner by drawing upon implicit differentiation. 





Another line of research partitions the parameter space into task-agnostic and  task-specific parameters. The latter are learned explicitly for each task, while the former remain fixed.
Bertinetto~\etal~\cite{luca} utilize a closed-form solution of the base learner, formulated as a ridge regression problem.
%
More recently, MetaOptNet~\cite{metaoptnet} employs a Support Vector Machine (SVM) as the task-specific model. Gradients are back-propagated through the SVM learning during meta-training using implicit differentiation of the optimality conditions of the convex problem.
While achieving promising performance, both these works~\cite{metaoptnet,luca} are restricted to specific types of task-specific models. 
In contrast, our approach allows for more general and flexible learning objectives. This enables us to, for instance, integrate entropy-based transductive learning during the meta-training stage. Moreover, in contrast to previous optimization-based few-shot classification methods, we propose to learn aspects of the task-specific learning formulation itself.


\section{Method}
\label{sec:method}


\begin{figure}[!t]
  \centering%
  \includegraphics*[width=1.0\linewidth]{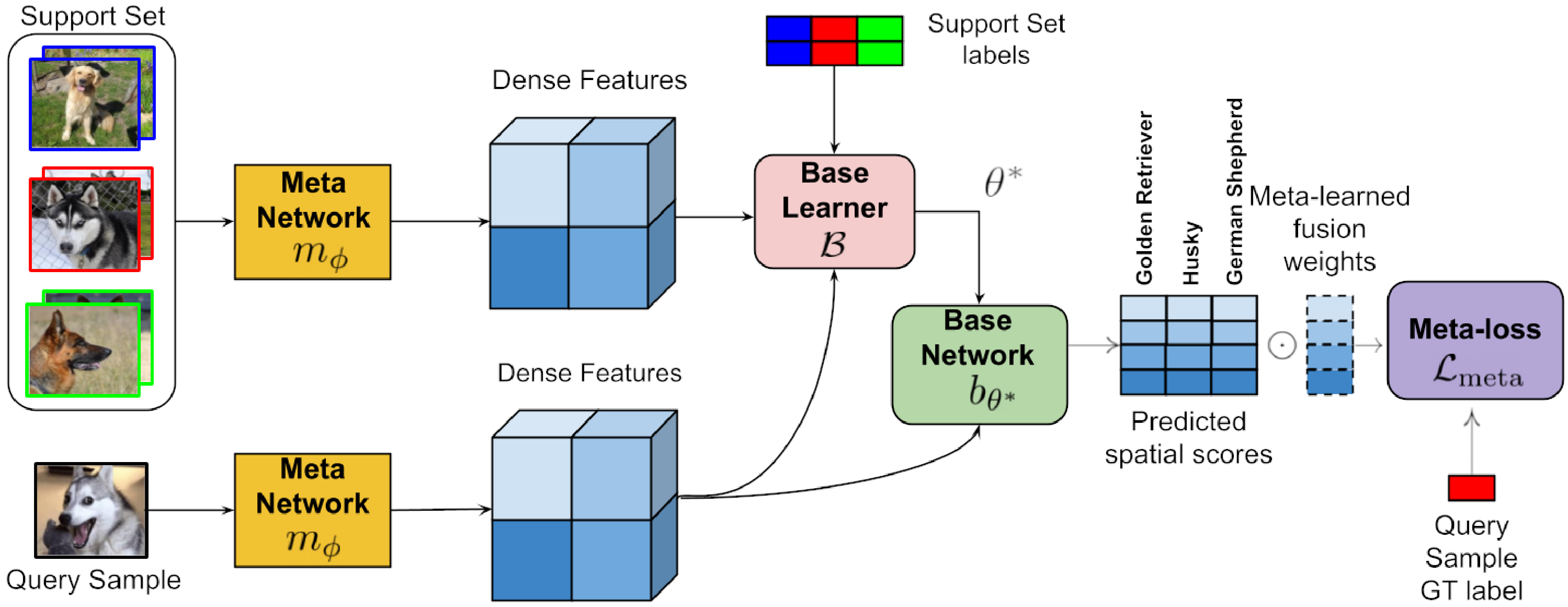}\vspace{-3mm}
  \caption{Overview of FIML (Few-Iteration Meta-Learning) for a 3-way 2-shot classification task. 
  }\vspace{-5mm}
  \label{fig:method}
\end{figure}

\subsection{Formulation}



Our meta-learning approach follows the paradigm in~\cite{metaoptnet,luca} by having two components: the base learner and a meta-learner. The aim of the base learner is to learn to perform a new task. It is, therefore, task-specific. The meta-learner, on the other hand, generalizes over a vast number of tasks. The main goal of the meta-learner is to learn a representation that is used by the task-specific base learner.
In a few-shot classification setting, a task $T=\{S, Q\}$ is characterized by a set of unlabeled \emph{query} images $Q = \{\tilde{x}_j, \tilde{y}_j\}_{j=1}^{\tilde{N}}$ which need to be classified, given only a few labeled image exemplars, called the support set $S = \{x_j, y_j\}_{j=1}^{N}$.
Here $x_j$ and $\tilde{x}_j$ are the $j^\text{th}$ image samples in the support and query sets, respectively. The corresponding labels in the support and query sets are denoted by $y_j$ and $\tilde{y}_j$, respectively. The support set contains $N=k\times n$ image-label pairs, \ie $n$ examples for each of the $k$ classes. Such a setting is called a $k$-way, $n$-shot classification problem. The query set $Q$ contains $\tilde{N}$ image-label pairs with the images sampled from the same class set as the support samples. For a given task, the support and query labels share the same label space $L=\{1,2,\dots, k\}$. The task support samples and the query samples for task $T$ are sampled from a data split $D$ of dataset $\mathcal{D}$. Here, $D\in\{D_\text{train}, D_\text{val}, D_\text{test}\}$ and,  $D_\text{train}$, $D_\text{val}$ and $D_\text{test}$ are the train, validation and test splits of $\mathcal{D}$. The dataset $\mathcal{D}$ also defines the domain of the few-shot classification problem.  The train, validation and test data splits are mutually disjoint and contain no common samples. Further, to  measure the meta network's generalization to unseen categories, the class sets for the train, validation and test splits ($C_\text{train}$, $C_\text{val}$ and $C_\text{test}$ respectively) are chosen to be pairwise disjoint.



The aim of the base learner $\mathcal{B}$ is to learn the optimal task specific base network $b_{\theta^*}$ conditioned on the meta network $m_\phi$. The general form of the base learner $\mathcal{B}$ is given by,
\begin{equation}
\label{eq:base_general}
    \mathcal{B}(T; m_\phi) \coloneqq \theta^* = \argmin_{\theta}\mathcal{L}_\text{base}(T, b_{\theta}; m_\phi), 
\end{equation}
where $\mathcal{L}_\text{base}$ is the base leaner loss. The prediction on the $j^{th}$ query sample  for task $T$ is then given by $b_{\theta^*}(m_\phi(\tilde{x}_j))$. In contrast to the task-specific base learner, the meta learner $\mathcal{M}$ trains a general model aiming to perform well on many, if not all tasks. $\mathcal{M}$ learns a generalized meta network $m_\phi^*$ over $T$. The meta-learning phase can be written as, 
\begin{equation}
    \mathcal{M}(D) \coloneqq \phi^* = \argmin_{\phi}\mathbb{E}_{T\sim \mathcal{T}}\mathcal{L}_\text{meta}(T, m_\phi; b_{\theta^*}), 
\end{equation}
where $\mathcal{T}$ is a set of $k$-way, $n$-shot classification problems sampled from $D$ and $\mathcal{L}_\text{meta}$ is the meta loss. The aim for the meta learner $\mathcal{M}$ is thus to learn a rich feature representation $m_{\phi^*}$ over which the base learner $\mathcal{B}$ can learn to classify the query samples. The training process mimics what happens during inference, except that the meta network $m_{\phi^*}$ remains fixed and only the base network $b_\theta$ is learned according to the given task $T$ shown to the base learner $\mathcal{B}$. Our base learner consists of an inductive objective, a transductive objective, and an optimization method, presented in the following sections. Fig. \ref{fig:method} visualizes our approach.

\subsection{Inductive Objective}








\label{sec:base_learner}

In this section, we describe the task-specific base network $b_{\theta}$ and the base learner $\mathcal{B}$. Given a task $T$, our aim is to learn a classifier $b_{\theta^{\*}}$, taking the representation learned by the meta-network as input. For an image $x$, the base network predicts classification scores $b_{\theta}(m_\phi(x)) \in \mathbb{R}^k$ for each of the $k$ classes. The base learner $\mathcal{B}$ minimizes $\mathcal{L}_\text{base}$ \wrt the parameters $\theta$. To this end, we start from a least squares objective as it allows effective optimization, which is crucial for practical meta-learning. 
%
%
%
In general, we consider a base learner objective of the following form,
\begin{equation}
\label{eq:loss_base}
    \mathcal{L}_\text{{ind}}(\theta) = \sum_{j=1}^{N}\left\|r\big(b_\theta(m_\phi(x_j)), \delta_{y_j}\big)\right\|^2  \,.
\end{equation}
Here, $r$ is a general residual error function and $\delta_{y_j}\in \mathbb{R}^k$ is a one-hot representation of the label $y_j$. 

The most straightforward choice for the residual function is to simply take the difference $r(s, \delta_{y}) = s - \delta_{y}$. For a linear base network $b_\theta$, this leads to a linear least squares ridge regression problem, as previously considered in a meta-learning context ~\cite{luca}.
However, such a loss is not a robust choice in the classification setting, since it penalizes any deviation from the target value $\delta_y$. In particular, \emph{easy} samples, residing on the correct side of the classification boundary, often dominate over \emph{hard} samples.
This issue is classically addressed by the SVM model, adopted by MetaOptNet~\cite{metaoptnet}, employing the hinge loss to ignore easy samples.


We take inspiration from the hinge loss, but propose a learnable parametrized alternative, suitable for the least-squares setting \eqref{eq:loss_base}.
Our residual function is defined as,
\begin{equation}
\label{eq:residual}
    r(s_j, \delta_{y_j}) = \max(z_j\cdot(l_j-s_j), a_j\cdot z_j\cdot(l_j-s_j)),
\end{equation}
where $z_j = 2\delta_{y_j} - 1$ signifies whether the class is positive or negative and $l_j=l_+\delta_{y_j} + l_-(1-\delta_{y_j})$ is the modified ground truth regression target with $l_+$ and $l_-$ being the target regression scores for the positive and negative classes (one-vs-all classification), respectively. Further, the predicted scores $s_j=b_\theta(x_j)$. For the final predictions, we apply a softmax on $s_j$. Intuitively, the parameters $l_+$ and $l_-$ define the margins of the classifier for the positive and the negative classes, respectively. The parameter $a_j = a_+\delta_{y_j} + a_-(1-\delta_{y_j})$, where $a_+$ and $a_-$ are the coefficient of leakage for the positive and negative classes respectively. The coefficients of leakage ($a_+$ and $a_-$) and the target regression scores ($l_+$ and $l_-$) are the free parameters in our base loss formulation which makes our loss more adaptive and robust at the same time. The free parameters of the loss are learned by the meta learner $\mathcal{M}$ during meta-training. Lastly, we note that although the $\max$ operation in \eqref{eq:residual} is not continuously differentiable, there are smooth alternatives, \eg the $\text{log}\text{-sum}\text{-exp}$ function, that well approximates its behaviour. 





\subsection{Transductive Objective}
Base learner modules for few-shot learning algorithms often suffer from high variance due to the low amount of labeled data available. However, the base learner $\mathcal{B}$ also has access to the unlabeled query samples, often ignored by meta-learning methods. These can aid the base learner by constraining the hypothesis search space.
This setting is very similar to a semi-supervised learning setup~\cite{lee2013pseudo_semi, oliver2018realistic_semi}. Recent metric-learning based few-shot classification methods~\cite{hou2019cross_can,liu2018learning_propogation} have demonstrated encouraging results by integrating transductive strategies. 
In contrast to previous works, we propose a strategy for integrating the transductive learning into the base learner itself. This imposes significant challenges since transductive objectives, such as the Shannon entropy, are non-convex in general. We tackle this problem by performing efficient convex approximations, allowing us to effectively combine transductive and inductive terms in our optimization-based meta-learning framework.


Although the base learner does not know to which class a query sample belongs, there is an important piece of information that it can exploit. Namely, that the query sample belongs to exactly \emph{one} of the categories in the task. This introduces a constraint, that can be formulated as an objective. In this work, we penalize the Shannon Entropy of the predictions on the query samples, stimulating the base learner to find classification parameters that yield confident predictions on the query set. 
Our transductive term is given by $\mathcal{L}_\text{tran}(\theta) = \sum_{j=1}^{\tilde{N}} H(\tilde{p}_j)$, where
\begin{equation}
\label{eq:trans-term}
    H(\tilde{p}_j) = -\sum_{c=1}^{k}\tilde{p}^c_j\log\tilde{p}^c_j= \log \sum_{c=1}^{k} e^{\tilde{s}_{j}^{c}} - \frac{\sum_{c=1}^{k}\tilde{s}_{j}^{c}e^{\tilde{s}_{j}^{c}}}{\sum_{c=1}^{k}e^{\tilde{s}_{j}^{c}}}.
\end{equation}
Here, $\tilde{s}_{j} = \beta b_\theta(m_\phi(\tilde{x}_j))$, $\tilde{x}_j$ is the  $j^{th}$ query sample and $\tilde{s}_{j}^{c}$ is its logit for class $c$. The second equality simply follows from the substitution of $\tilde{p}_j = \text{SoftMax}(\tilde{s}_j)$. Note that we have introduced a temperature scaling parameter $\beta$. By learning this parameter during meta-training, our approach learns to calibrate the query classification probabilities in order to benefit the transductive learning. 

Our final base learner objective, integrating both the inductive \eqref{eq:loss_base} and the transductive \eqref{eq:trans-term} terms is thus,
\begin{equation}
\label{eq:base-loss}
    \mathcal{L}_\text{base}(\theta) = \mathcal{L}_\text{ind}(\theta) + \lambda_\text{tran} \mathcal{L}_\text{tran}(\theta) + \lambda_\text{reg} \|\theta\|^2 \,.
\end{equation}
The second term is a regularization on the base learned parameters $\theta$. Since the objective \eqref{eq:base-loss} is utilized in the base learner, we can even learn the importance weights $\lambda_\text{tran}$ and $\lambda_\text{reg}$ during meta-training, thus circumventing the need for tuning by hand. Next, we derive our base learner $\theta^* = \mathcal{B}(T; m_\phi)$ by applying an unrolled optimization procedure to minimize \eqref{eq:base-loss}. Note that, in contrast to semi-supervised methods, our approach does not utilize any additional unlabeled data during training. The transductive term instead allows us to better exploit the query sample during inference.

\subsection{Base Learner Optimization}
\label{sec:base_init}


Our base learner \eqref{eq:base_general} is implemented by applying an iterative optimization algorithm to our objective \eqref{eq:base-loss}. 
%
Meta-learning methods popularly resort to the standard gradient descent algorithm~\cite{finn2017maml}. However, we primarily consider a linear base model $b_\theta(x) = \theta x$, where $\theta$ are the linear classification weights, allowing more effective strategies that enjoy substantially faster convergence to be employed. We therefore adapt the Steepest Descent based strategy~\cite{dimp} to our setting. 


We start from a positive definite quadratic approximation of the objective \eqref{eq:base-loss}. Although our objective is non-convex, such an approximation can be derived using Gauss-Newton for \eqref{eq:loss_base} and a positive definite approximate Hessian of \eqref{eq:trans-term}. The quadratic approximation provides the optimal step length $\alpha^{(d)}$ in the gradient direction through a simple closed-form expression. The optimization iteration is expressed as,
\begin{equation}
\label{eq:optim}
\begin{split}
    &\theta^{(d+1)} = \theta^{(d)} - \alpha^{(d)} \nabla\mathcal{L}_\text{base}(\theta^{(d)}) \text{ where,}\\ &\alpha^{(d)} = \frac{\nabla\mathcal{L}_\text{base}(\theta^{(d)})^T\nabla\mathcal{L}_\text{base}(\theta^{(d)})}{\nabla\mathcal{L}_\text{base}(\theta^{(d)})^TH^{(d)}\nabla\mathcal{L}_\text{base}(\theta^{(d)})} \,.
\end{split}
\end{equation}
Here, $d$ denotes the iteration number. Note that the positive definite Hessian approximation $H^{(d)}$ does not need to be computed explicitly. Instead, both the computation of the gradient $\nabla\mathcal{L}_\text{base}$ and the Hessian-gradient product $H^{(d)} \nabla\mathcal{L}_\text{base}$ can be efficiently implemented using standard network operations or double-backpropagation.

To further reduce the number of required iterations in \eqref{eq:optim}, we propose an effective initialization strategy to obtain $\theta^{(0)}$. We express the initial classification weights $\theta^{(0)}_c$ of class $c$ as a linear combination between the average positive $f^c_\text{pos}$ and negative $f^c_\text{neg}$ feature vectors in the support set, $\theta^{(0)}_c = \kappa^c f^c_\text{pos} - \tau^c f^c_\text{neg}$ where, $f^c_\text{pos}=\frac{1}{n}\sum_{j=1}^N\mathbbm{1}_{y_j=c}m_\phi(x_j)$ and, $f^c_\text{neg}=\frac{1}{N-n}\sum_{j=1}^N\mathbbm{1}_{y_j\neq c}m_\phi(x_j)$. Here, $N$ is the number of samples in the support set $S$ for task $T$ for an $n$-shot problem. The two scalars $\kappa^c, \tau^c \in \mathbb{R}$ are found by defining two linear constraints $(f^c_\text{pos})^T\theta^{(0)}_c = o_+$ and $(f^c_\text{neg})^T\theta^{(0)}_c = o_-$, where $o_+,o_-$ represent the classification score for $f^c_\text{pos}$ and $f^c_\text{neg}$. The closed-form expression of $\tau^c,\kappa^c$ is easily obtained by solving the 2-dimensional linear system. We learn the values of $o_+,o_-$ during meta-training. Unlike Proto-MAML \cite{Triantafillou2020Meta-Dataset:} which initializes the task-specific linear layer with the Prototypical Network-equivalent weights, our base learner initializer also aims to exploit the negative examples which increases the discriminative ability.

\subsection{Dense Classification}

\label{sec:dense}

In this section, we further address the scarcity of labeled data by integrating a dense classification strategy, utilizing samples extracted from different spatial locations in the image. 
Recently, Lifchitz~\etal~\cite{lifchitz2019dense} demonstrated the benefit of using dense features for few-shot classification tasks. Standard deep embedding networks $m_\phi$ terminate with a global average pooling layer. This leads to a loss of information that could prove detrimental to the performance of few-shot classification algorithms. We, therefore, utilize dense spatial features before the global average pooling layer in $m_\phi$. Let $m_\phi^l(x)$ be the feature vector at spatial index $l$. Our previously described base learner is  modified to operate on these localized features by simply treating them as individual examples in the support and query sets, respectively. Both, our inductive \eqref{eq:loss_base} and transductive \eqref{eq:trans-term} objectives thus includes one term per spatial location in the dense feature map obtained from $m_\phi$. 
%
%
While this strategy allows us to learn from multiple localized examples, our task is to generate one final prediction per query image. This is achieved through a spatial fusion procedure,
\begin{equation}
\label{eq:querz_pred}
    \tilde{s}_j = \sum_{l} v_l b_{\theta^*}\left(m_\phi^l(\tilde{x}_j)\right) \,,
\end{equation}
where $\{v_l\}_{l}$ is a set of spatial weights that are learned during meta-training. These weights signify how much emphasis must be given to a prediction of the base network $b_{\theta^*}$ for a certain spatial location. 

The fused scores \eqref{eq:querz_pred} serve as the logits for the final classification output. During meta-training, we minimize the cross-entropy to the ground-truth probability vector $\hat{p}_j$ for each query sample $\tilde{x}_j$ in the task as,
\begin{equation}
\label{eq:meta-loss}
    \mathcal{L}_\text{meta}(\phi, \psi) = -\frac{1}{\tilde{N}}\sum_{j=1}^{\tilde{N}}\sum_{c=1}^{k}\hat{p}^c_j\log\tilde{p}^c_j.  
\end{equation}
Here $\tilde{p}_j=\text{SoftMax}(\tilde{s}_j)$. In addition to the parameters $\phi$ of the embedding network, we meta-learn parameters of our base-learner $\psi = \{l_+,l_-,a_+,a_-,\beta,\lambda_\text{tran},\lambda_\text{reg},o_+,o_-,v_l\}$. The latter include parameters of our inductive and transductive objectives, along with the optimization parameters and the fusion weights. 

\section{Experiments}



\subsection{Implementation Details}
We implement our approach using PyTorch~\cite{paszke2017pytorch}. Our experiments are conducted with two different embedding networks $m_\phi$ commonly used for few-shot classification, namely ResNet-12 and WideResNet-28-10. 
Both the backbones are trained from scratch during meta-training. We follow the meta-training and meta-testing strategies used in~\cite{metaoptnet}. 
As an optimizer, we use SGD~\cite{sgd} with Nesterov momentum of $0.9$ and weight decay of $0.0005$. We meta-train the model for 60 epochs, each containing 1000 batches. Each batch consists of 16 tasks.
The number of iterations for base network optimizer at train time  was set to 10 during meta-training and 15 during meta-testing. 
At training time, the best model is chosen based on 5-way, $n$-shot classification accuracy on the validation set. 
Following~\cite{metaoptnet}, the network is first trained using a 15-shot setting. Since we aim to learn the base learner itself for a particular distribution of tasks, we further fine-tune the base learner parameters $\psi$ in \eqref{eq:meta-loss} for 10 epochs for the specific shot setting using a learning rate of $0.001$.
 Now, we give a brief overview of the few-shot classification benchmarks, which we use for evaluating our approach.

\subsection{Few-shot Classification Benchmarks}
\textbf{ImageNet Derivatives:} The miniImageNet ~\cite{ravi2016optimization} and the tieredImageNet~\cite{ren2018tieredimagenet} few-shot classification benchmarks are derived from ILSVRC-2012~\cite{russakovsky2015imagenet}. The image size in both the benchmarks is 84$\times$84.  Conversely, to minimize semantic similarity between the splits, tieredImageNet splits data based on the superclasses.

\textbf{CIFAR-100 Derivatives:} Both, the CIFAR-FS~\cite{luca} and the FC100~\cite{oreshkin2018tadam} benchmarks encompass the full CIFAR-100 dataset~\cite{krizhevsky2009cifar}. While the CIFAR-100 dataset is split based on the subclasses to derive the CIFAR-FS dataset; FC100, similar to tieredImageNet, splits CIFAR-100 based on superclasses to minimize semantic similarity. The image size is 32$\times$32 in both the datasets.     

\subsection{Ablation Study}
We perform an ablative study on the two larger few-shot classification datasets, namely the miniImageNet and tieredImageNet. The methods are compared in terms of the few-shot classification accuracy (\%) with a 95\% confidence interval. Results are reported in Tab.~\ref{tab:ablation1}. 
\begin{table}[!t]
 \centering%
 \caption{Ablative study of our approach on tieredImageNet and miniImageNet datasets. Results are reported in terms of accuracy (\%) with 95\% confidence interval.}\vspace{-2 mm}
	\resizebox{1.0\columnwidth}{!}{%
 \begin{tabular}{lcccc}
	\toprule
	&\multicolumn{2}{c}{\textbf{miniImageNet 5-way}}&\multicolumn{2}{c}{\textbf{tieredImageNet 5-way}}\\
 \cline{2-3} \cline{4-5}
	&\textbf{1-shot}&\textbf{5-shot}&\textbf{1-shot}&\textbf{5-shot}\\\midrule
	\textbf{Baseline}&64.73$\pm$0.73&80.89$\pm$0.54&61.97$\pm$0.69&78.12$\pm$0.48\\
	\textbf{+Initializer}&65.32$\pm$0.71&81.06$\pm$0.57&62.13$\pm$0.68&78.56$\pm$0.45\\
	\textbf{+Dense Features}&66.41$\pm$0.64&82.96$\pm$0.51&63.01$\pm$0.64&79.45$\pm$0.44\\
	\textbf{+LearnLoss}&67.27$\pm$0.66&83.83$\pm$0.52&63.67$\pm$0.65&80.17$\pm$0.46\\
	\textbf{+Transductive}&69.92$\pm$0.64&84.41$\pm$0.55&65.00$\pm$0.64&80.52$\pm$0.43\\
 \bottomrule
\end{tabular}}\vspace{-7.5mm}
 \label{tab:ablation1}%
\end{table}

\textbf{Baseline:} As the baseline, we meta-train and meta-test our framework with fixed hyperparameters ($l_+=1$, $l_-=-1$, $a_+=1$, $a_-=1$ in \ref{eq:residual} and $\lambda_\text{reg}=0.01$ in \eqref{eq:base-loss}). Note that this configuration implies a linear ridge regression objective, similar to \cite{luca}. Further we use a zero initializer $\theta^{(0)} = 0$ in our base learner (Sec.~\ref{sec:base_init}). We do not include the transductive objective ($\lambda_\text{tran}=0$) and do not use the dense classification strategy described in Sec.~\ref{sec:dense}. 

\textbf{+Initializer:} This version adds the support set based initialization of $\theta^{(0)}$ discussed in section \ref{sec:base_init} to the baseline. The parameters $o_+$ and $o_-$ are fixed to the fixed value $1$. \textbf{+Initializer} improves over the baseline with a relative gain of 0.9\% for 1-shot evaluation on the tieredImageNet dataset. This ablation suggests that our base network initializer leads to a faster convergence for the base learner.  

\textbf{+DenseFeatures:} Next, we investigate the effect of using the spatial dense classification strategy, discussed in Sec.~\ref{sec:dense}, by adding it to the previous configuration. In this version, we keep the spatial fusion weights fixed along with the other base learner parameters. Specifically, we use an average pooling of logits in \eqref{eq:querz_pred} by setting $v_l = 1/\textsc{L}$, where $\textsc{L}$ is the number of spatial feature locations. The utilization of dense features leads to a significant increase in the performance over +IntelInit, with relative gains of 1.7\% and 2.3\% for 1-shot and 5-shot performance, respectively, on tieredImageNet. 


\textbf{+LearnLoss:} Here, we additionally learn all the parameters $\psi$ of the base learner, defined in Sec.~\ref{sec:dense}. These include the parameters of the inductive loss, optimizer, and the spatial fusion weights for dense classification. By exploiting the flexibility of our framework to learn the optimal base learner through meta-training, \textbf{+LearnLoss} achieves a substantial relative increment of 1.3\% and 1.0\% for 1-shot and 5-shot performance on tieredimageNet. Results on miniImageNet follow a similar trend, with relative improvements of 1\% and 0.9\% for 1-shot and 5-shot respectively. 

\textbf{+Transductive:} Finally, we add the transductive term \ref{eq:trans-term} to our overall objective \eqref{eq:base-loss}. The free parameters of the transductive term, \ie the importance weight $\lambda_\text{tran}$ and the temperature scaling $\beta$ are learned by the meta-learner. Adding the transductive loss for learning an optimal base network gives a major improvement in 1-shot performance, with relative gains of  3.9\% and 2.1\% for tieredimageNet and miniImageNet, respectively. The improvement in the 5-shot setting is more modest, but still significant. The results follow an expected trend, since the potential benefit of transductive learning increases when the number of labeled examples are reduced. Interestingly, a more detailed investigation showed that the learned importance weight $\lambda_\text{trans}$ for the transductive objective was substantially larger in the 1-shot scenario compared to when trained for the 5-shot setting. This demonstrates that our approach is capable of meta-learning not only the feature embedding, but also the base learner itself. We use this version of our approach for the state-of-the-art comparisons presented in Sec. \ref{sec:sota}.

\begin{table}[!t]
 \centering%
 \caption{Cross-validation across datasets (WRN-28-10 backbone) for a 5-way task.}\vspace{-2 mm}
 \label{tab:overfit}%
	\resizebox{\linewidth}{!}{%
 \begin{tabular}{llcc}
	\toprule
	&&\textbf{FIML}&\textbf{FIML-CrossVal}\\\midrule
	\multirow{2}{*}{\textbf{tieredImageNet}}&1-shot&72.97$\pm$0.47&73.06$\pm$0.50\\
 &5-shot&86.12$\pm$0.37&85.90$\pm$0.41\\
 \midrule
 \multirow{2}{*}{\textbf{miniImageNet}}&1-shot&67.89$\pm$0.42&67.95$\pm$0.39\\
 &5-shot&82.31$\pm$0.33&82.46$\pm$0.35\\
 \midrule
 	\multirow{2}{*}{\textbf{FC100}}&1-shot&45.01$\pm$0.46&44.91$\pm$0.44\\
 &5-shot&58.96$\pm$0.51&59.27$\pm$0.54\\
 \midrule
 \multirow{2}{*}{\textbf{CIFAR-FS}}&1-shot&77.21$\pm$0.46&77.08$\pm$0.41\\
 &5-shot&88.49$\pm$0.33&88.70$\pm$0.36\\
 \bottomrule
\end{tabular}}\vspace{-7.5mm}
\end{table}

\textbf{Cross-validation across datasets:}  Learning the loss parameters $\psi$ on a single dataset can lead to overfitting. To analyze potential dataset-specific overfitting, we perform cross-validation of the learned loss parameters $\psi$ across datasets (Tab.~\ref{tab:overfit}). For reporting the cross-validation accuracies (FIML-CrossVal), we evaluate the model trained for tieredImageNet on FC100 and CIFAR-100. Similarly, we evaluate the $\psi$ parameters trained for FC100 to report accuracies on the datasets derived from miniImageNet. When using the loss parameters $\psi$ learned on tieredImageNet, the average classification accuracy only changes by 0.10\% and 0.31\% for 1-shot and 5-shot, respectively on the FC100 dataset, while the opposite leads to a change of 0.09\% and 0.22\% on the tieredImageNet dataset, respectively. The learned loss thus generalizes well to other datasets yielding comparable performances to FIML in all cases. 


\begin{table*}[!htbp]
\centering
\caption{State-of-the-art: Few-shot classification accuracy (\%) with 95\% confidence intervals on meta-test splits. 4-layer convolutional networks are represented as a sequence denoting the number of filters in each layer. The red and blue colours signify the best and the second-best methods for a particular backbone, respectively. Further, - means that the results have not been reported for that method.
}\vspace{-2mm}
\resizebox{1.0\linewidth}{!}{%
\begin{tabular}{llcccccccc}
 \toprule
 &&\multicolumn{2}{c}{\textbf{miniImageNet 5-way}}&\multicolumn{2}{c}{\textbf{tieredImageNet 5-way}}&\multicolumn{2}{c}{\textbf{CIFAR-FS 5-way}}&\multicolumn{2}{c}{\textbf{FC100 5-way}}\\
 \cline{3-4} \cline{5-6} \cline{7-8} \cline{9-10}
 \textbf{Method}&\textbf{Backbone}&\textbf{1-shot}&\textbf{5-shot}&\textbf{1-shot}&\textbf{5-shot}&\textbf{1-shot}&\textbf{5-shot}&\textbf{1-shot}&\textbf{5-shot}\\
 \midrule
 CAVIA \cite{zintgraf2018cavia}&32-32-32-32&47.24$\pm$0.65&59.05$\pm$0.54&-&-&-&-&-&-\\
 MAML \cite{finn2017maml}&32-32-32-32&48.70$\pm$1.84&63.11$\pm$0.92& 51.67$\pm$1.81&70.30$\pm$1.75&58.90$\pm$1.90&71.50$\pm$1.00&-&-\\
 Matching Networks \cite{vinyals2016matching}&64-64-64-64&43.56$\pm$0.84&55.31$\pm$0.73&-&-&-&-&-&-\\
 Relation Networks\cite{sung2018learning}&64-96-128-256&50.44$\pm$0.82&65.32$\pm$0.70&54.48$\pm$0.93&71.32$\pm$0.78&55.00$\pm$1.00&69.30$\pm$0.80&-&-\\
 LSTM Meta Learner \cite{ravi2016optimization}&64-64-64-64&43.44$\pm$0.77&60.60$\pm$0.71&-&-&-&-&-&-\\
 Transductive Propagation\cite{liu2018learning_propogation}&64-64-64-64&55.51$\pm$0.86&69.86$\pm$0.65&59.91$\pm$0.49&73.30$\pm$0.45&-&-&-&-\\
 R2D2 \cite{luca}&96-192-384-512&51.20$\pm$0.60&68.80$\pm$0.10&-&-&65.30$\pm$0.20&79.40$\pm$0.10&-&-\\
 \midrule
 TADAM \cite{oreshkin2018tadam}&ResNet-12&58.50$\pm$0.30&76.70$\pm$0.30&-&-&-&-&40.10$\pm$0.40&56.10$\pm$0.40\\
 MetaOptNet-SVM \cite{metaoptnet}&ResNet-12&62.64$\pm$0.61&78.63$\pm$0.46&65.99$\pm$0.72&81.56$\pm$0.53&72.00$\pm$0.70&\textcolor{blue}{\textbf{84.20$\pm$0.50}}&41.10$\pm$0.60&55.50$\pm$0.60\\
 MetaOptNet-SVM\cite{metaoptnet}+Dense&ResNet-12&61.69$\pm$0.55&\textcolor{blue}{\textbf{79.13$\pm$0.43}}&65.17$\pm$0.67&81.44$\pm$0.45&\textcolor{blue}{\textbf{72.23$\pm$0.65}}&83.65$\pm$0.47&41.56$\pm$0.59&55.97$\pm$0.55\\
 DC \cite{liu2018learning_propogation}&ResNet-12&61.26$\pm$0.20& 79.01$\pm$0.13&-&-&-&-&\textcolor{blue}{\textbf{42.04$\pm$0.17}}&\textcolor{blue}{\textbf{57.05$\pm$0.16}}\\
 CAN \cite{hou2019cross_can}&ResNet-12&63.85$\pm$0.48&\textcolor{blue}{\textbf{79.44$\pm$0.34}}&\textcolor{blue}{\textbf{69.89$\pm$0.51}}&\textcolor{blue}{\textbf{84.23$\pm$0.37}}&-&-&-&-\\
 CAN+Transduction \cite{hou2019cross_can}&ResNet-12&\textcolor{red}{\textbf{67.19$\pm$0.55}}&\textcolor{red}{\textbf{80.64$\pm$0.35}}&\textcolor{red}{\textbf{73.21$\pm$0.58}}&\textcolor{red}{\textbf{84.93$\pm$0.38}}&-&-&-&-\\
 \textbf{FIML (Ours)}&ResNet-12&\textcolor{blue}{\textbf{65.00$\pm$0.64}}&\textcolor{red}{\textbf{80.52$\pm$0.43}}&\textcolor{blue}{\textbf{69.92$\pm$0.64}}&\textcolor{blue}{\textbf{84.41$\pm$0.55}}&\textcolor{red}{\textbf{75.13$\pm$0.66}}&\textcolor{red}{\textbf{85.61$\pm$0.44}}&\textcolor{red}{\textbf{42.75$\pm$0.56}}&\textcolor{red}{\textbf{57.23$\pm$0.54}}\\
 \midrule
 LEO \cite{leo}&WideResNet-28-10&61.76$\pm$0.08&77.59$\pm$0.12&66.33$\pm$0.05&81.44$\pm$0.09&-&-&-&-\\
 Trans-FT \cite{dhillon2019baseline} &WideResNet-28-10&\textcolor{blue}{\textbf{65.73$\pm$0.78}}&\textcolor{blue}{\textbf{78.40$\pm$0.52}}&\textcolor{red}{\textbf{73.34$\pm$0.71}}&\textcolor{blue}{\textbf{85.50$\pm$0.50}}&76.58$\pm$0.68&85.79$\pm$0.50&43.16$\pm$0.59&57.57$\pm$0.55\\
 \textbf{FIML (Ours)}&WideResNet-28-10&\textcolor{red}{\textbf{67.89$\pm$0.42}}&\textcolor{red}{\textbf{82.31$\pm$0.33}}&\textcolor{blue}{\textbf{72.97$\pm$0.47}}&\textcolor{red}{\textbf{86.12$\pm$0.37}}&\textcolor{red}{\textbf{77.21$\pm$0.46}}&\textcolor{red}{\textbf{88.49$\pm$0.33}}&\textcolor{red}{\textbf{45.01$\pm$0.46}}&\textcolor{red}{\textbf{58.96$\pm$0.51}}\\
 \bottomrule
\end{tabular}}\vspace{-5mm}
\label{tab:big_datasets}
\end{table*}

\subsection{State-of-the-Art}
\label{sec:sota}
We compare our approach FIML with state-of-the-art methods for few-shot classification. In Tab.~\ref{tab:big_datasets} we report results on the datasets derived from ImageNet and CIFAR-100, respectively. Among the compared methods, R2D2~\cite{luca} and MetaOptNet-SVM~\cite{metaoptnet} employ an optimization-based base learner that predicts the parameters of a linear classification head. Our approach significantly outperforms these methods. Notably, when employing the same ResNet-12 backbone, our approach achieves relative improvements of 3.6\% and 2.4\% for 1-shot and 5-shot performance, respectively, on the larger tieredImageNet dataset. Moreover, compared to MetaOptNet-SVM~\cite{metaoptnet} and R2D2~\cite{luca}, our framework can utilize a wider class of objective functions, allowing us to integrate the non-convex transductive objective \eqref{eq:trans-term} and also to meta-learn important parameters of the objective and the base learner itself. We also implemented a dense classification version of MetaOptNet-SVM (MetaOptNet-SVM+Dense), but did not find it to yield any significant improvement. We also compare the computational time of FIML and MetaOptNet+Dense on tieredImageNet. For MetaOptNet+Dense, the number of QP solver iterations were carefully set based on the validation set for the best test accuracy, whereas the standard setting of 15 iterations were run for the base learner optimizer to time FIML. Tab.~\ref{tab:time} shows the timings in milliseconds (\textit{ms}) per episode for 1-shot, 5-shot and 15-shot tasks. It is evident that FIML scales better than MetaOptNet+Dense \wrt the number of shots during inference. 

\begin{table}
\centering
\caption{Comparision of mean inference times (in \textit{ms}) with 95\% confidence interval of FIML with MetaOptNet-SVM (with dense features) and Trans-FT on the tieredImageNet dataset for a 5-way task.}\vspace{-2mm}
	\resizebox{1.0\linewidth}{!}{%
\begin{tabular}{llccc}
	\toprule
	&\textbf{Backbone}&\textbf{1-shot}&\textbf{5-shot}&\textbf{15-shot}
	\\\midrule
MetaOptNet-SVM\cite{metaoptnet}+Dense&ResNet-12&79$\pm$15&109$\pm$20&174$\pm$16\\
	FIML (ours)&ResNet-12&67$\pm$18&91$\pm$17&133$\pm$21\\\midrule
	Trans-FT \cite{dhillon2019baseline}&WideResNet-28-10&20800&-&-\\
	FIML (ours)&WideResNet-28-10&107$\pm$24&159$\pm$30&261$\pm$28\\\bottomrule
\end{tabular}}\vspace{-6mm}
\label{tab:time}
\end{table}

Among other compared methods, DC~\cite{liu2018learning_propogation} and CAN~\cite{hou2019cross_can} utilize dense classification. The latter is a recent metric-learning based method, employing pairwise cross attention maps for the class prototypes and the query samples for extracting discriminative features. Further, CAN+Transduction augment the support set with the query samples to learn more representative class prototypes. The query samples are assigned to a particular class based on the nearest neighbors of the query sample in the support set. This approach achieves strong performance, particularly for the 1-shot case. While our work is the first to integrate transductive strategies into an optimization-based meta-learning framework, the results of CAN show that there is scope for further improvements in this direction. When using the WideResNet backbone, our approach significantly improves over CAN+Transduction in the 5-shot setting, while achieving similar or better for 1-shot. 

Lastly, we compare our approach with the very recent method Trans-FT~\cite{dhillon2019baseline}. This approach does not perform meta-learning, but directly addresses the few-shot learning problem by learning a generalizable representation using the class labels in the training set. Given the learned representation, the few-shot classifier is trained using SGD during testing using both an inductive classification loss and a transductive loss. Trans-FT can therefore not be used in online scenarios, where inference time is critical. On tieredImageNet, our method improves on Trans-FT in the 5-shot case, with a slight degradation in 1-shot performance. However, our approach outperforms Trans-FT on all other datasets, achieving relative gains of 3.3\% and 5\% for 1-shot and 5-shot evaluations, respectively, on miniImageNet. Further, a major disadvantage of ~\cite{dhillon2019baseline} is its computational complexity. In fact, our approach is about 200$\times$ faster (Tab.~\ref{tab:ablation1}): $20.8$s is reported in ~\cite{dhillon2019baseline} while we achieve $0.107\pm0.024$s in the same setting (one 1-shot, 5-way, task with 15 queries) and backbone (WideResNet-28-10). Runtime is of crucial importance in applications involving learning agents and other time-critical systems. 
This indicates the advantage of meta-learning based methods. Further, in contrast to the manual hand-tuning required in~\cite{dhillon2019baseline}, our approach automatically learns the crucial hyper-parameters associated with the few-shot objective, including the temperature scaling and the balance between the losses. 

\section{Conclusion}
We propose an iterative optimization-based meta-learning method FIML, which integrates dense features with a novel adaptive fusion module in a few-shot setting. FIML consists of a base-learner, employing a robust classification loss. A transductive loss term is also integrated into our flexible framework, forcing the base network to make confident predictions on the query samples. Further, a support set based initialization of the linear base network aids the iterative unrolled optimizer in a faster convergence of thebase network. We experimentally validate our approach on four few-shot classification benchmarks and set a new state-of-the-art for optimization-based meta-learning.

\bibliography{main}

\begin{thebibliography}{10}

\bibitem{vinyals2016matching}
O.~Vinyals, C.~Blundell, T.~Lillicrap, D.~Wierstra, {\em et~al.}, ``Matching
  networks for one shot learning,'' in {\em Advances in neural information
  processing systems}, pp.~3630--3638, 2016.

\bibitem{snell2017prototypical}
J.~Snell, K.~Swersky, and R.~Zemel, ``Prototypical networks for few-shot
  learning,'' in {\em Advances in neural information processing systems},
  pp.~4077--4087, 2017.

\bibitem{finn2017maml}
C.~Finn, P.~Abbeel, and S.~Levine, ``Model-agnostic meta-learning for fast
  adaptation of deep networks,'' in {\em Proceedings of the 34th International
  Conference on Machine Learning-Volume 70}, pp.~1126--1135, JMLR. org, 2017.

\bibitem{closerlook}
W.~Chen, Y.~Liu, Z.~Kira, Y.~F. Wang, and J.~Huang, ``A closer look at few-shot
  classification,'' in {\em 7th International Conference on Learning
  Representations, {ICLR} 2019, New Orleans, LA, USA, May 6-9, 2019},
  OpenReview.net, 2019.

\bibitem{lake2015human}
B.~M. Lake, R.~Salakhutdinov, and J.~B. Tenenbaum, ``Human-level concept
  learning through probabilistic program induction,'' {\em Science}, vol.~350,
  no.~6266, pp.~1332--1338, 2015.

\bibitem{sung2018learning}
F.~Sung, Y.~Yang, L.~Zhang, T.~Xiang, P.~H. Torr, and T.~M. Hospedales,
  ``Learning to compare: Relation network for few-shot learning,'' in {\em
  Proceedings of the IEEE Conference on Computer Vision and Pattern
  Recognition}, pp.~1199--1208, 2018.

\bibitem{ravi2016optimization}
S.~Ravi and H.~Larochelle, ``Optimization as a model for few-shot learning,''
  in {\em 5th International Conference on Learning Representations, {ICLR}
  2017, Toulon, France, April 24-26, 2017, Conference Track Proceedings},
  OpenReview.net, 2017.

\bibitem{ImplicitGrad}
A.~Rajeswaran, C.~Finn, S.~M. Kakade, and S.~Levine, ``Meta-learning with
  implicit gradients,'' in {\em Advances in Neural Information Processing
  Systems 32: Annual Conference on Neural Information Processing Systems 2019,
  NeurIPS 2019, 8-14 December 2019, Vancouver, BC, Canada} (H.~M. Wallach,
  H.~Larochelle, A.~Beygelzimer, F.~d'Alch{\'{e}}{-}Buc, E.~B. Fox, and
  R.~Garnett, eds.), pp.~113--124, 2019.

\bibitem{metaoptnet}
K.~Lee, S.~Maji, A.~Ravichandran, and S.~Soatto, ``Meta-learning with
  differentiable convex optimization,'' in {\em Proceedings of the IEEE
  Conference on Computer Vision and Pattern Recognition}, pp.~10657--10665,
  2019.

\bibitem{luca}
L.~Bertinetto, J.~F. Henriques, P.~H.~S. Torr, and A.~Vedaldi, ``Meta-learning
  with differentiable closed-form solvers,'' in {\em 7th International
  Conference on Learning Representations, {ICLR} 2019, New Orleans, LA, USA,
  May 6-9, 2019}, OpenReview.net, 2019.

\bibitem{dhillon2019baseline}
G.~S. Dhillon, P.~Chaudhari, A.~Ravichandran, and S.~Soatto, ``A baseline for
  few-shot image classification,'' in {\em 8th International Conference on
  Learning Representations, {ICLR} 2020, Addis Ababa, Ethiopia, April 26-30,
  2020}, OpenReview.net, 2020.

\bibitem{koch2015siamese}
G.~Koch, R.~Zemel, and R.~Salakhutdinov, ``Siamese neural networks for one-shot
  image recognition,'' in {\em ICML deep learning workshop}, vol.~2, Lille,
  2015.

\bibitem{first_order_reptile}
A.~Nichol, J.~Achiam, and J.~Schulman, ``On first-order meta-learning
  algorithms,'' {\em CoRR}, vol.~abs/1803.02999, 2018.

\bibitem{zintgraf2018cavia}
L.~M. Zintgraf, K.~Shiarlis, V.~Kurin, K.~Hofmann, and S.~Whiteson, ``Fast
  context adaptation via meta-learning,'' in {\em Proceedings of the 36th
  International Conference on Machine Learning, {ICML} 2019, 9-15 June 2019,
  Long Beach, California, {USA}} (K.~Chaudhuri and R.~Salakhutdinov, eds.),
  vol.~97 of {\em Proceedings of Machine Learning Research}, pp.~7693--7702,
  {PMLR}, 2019.

\bibitem{lee2013pseudo_semi}
D.-H. Lee, ``Pseudo-label: The simple and efficient semi-supervised learning
  method for deep neural networks,'' in {\em Workshop on challenges in
  representation learning, ICML}, vol.~3, p.~2, 2013.

\bibitem{oliver2018realistic_semi}
A.~Oliver, A.~Odena, C.~A. Raffel, E.~D. Cubuk, and I.~Goodfellow, ``Realistic
  evaluation of deep semi-supervised learning algorithms,'' in {\em Advances in
  Neural Information Processing Systems}, pp.~3235--3246, 2018.

\bibitem{hou2019cross_can}
R.~Hou, H.~Chang, M.~Bingpeng, S.~Shan, and X.~Chen, ``Cross attention network
  for few-shot classification,'' in {\em Advances in Neural Information
  Processing Systems}, pp.~4005--4016, 2019.

\bibitem{liu2018learning_propogation}
Y.~Liu, J.~Lee, M.~Park, S.~Kim, E.~Yang, S.~J. Hwang, and Y.~Yang, ``Learning
  to propagate labels: Transductive propagation network for few-shot
  learning,'' in {\em 7th International Conference on Learning Representations,
  {ICLR} 2019, New Orleans, LA, USA, May 6-9, 2019}, OpenReview.net, 2019.

\bibitem{dimp}
G.~Bhat, M.~Danelljan, L.~V. Gool, and R.~Timofte, ``Learning discriminative
  model prediction for tracking,'' in {\em Proceedings of the IEEE
  International Conference on Computer Vision}, pp.~6182--6191, 2019.

\bibitem{Triantafillou2020Meta-Dataset:}
E.~Triantafillou, T.~Zhu, V.~Dumoulin, P.~Lamblin, U.~Evci, K.~Xu, R.~Goroshin,
  C.~Gelada, K.~Swersky, P.-A. Manzagol, and H.~Larochelle, ``Meta-dataset: A
  dataset of datasets for learning to learn from few examples,'' in {\em
  International Conference on Learning Representations}, 2020.

\bibitem{lifchitz2019dense}
Y.~Lifchitz, Y.~Avrithis, S.~Picard, and A.~Bursuc, ``Dense classification and
  implanting for few-shot learning,'' in {\em Proceedings of the IEEE
  Conference on Computer Vision and Pattern Recognition}, pp.~9258--9267, 2019.

\bibitem{paszke2017pytorch}
A.~Paszke, S.~Gross, S.~Chintala, G.~Chanan, E.~Yang, Z.~DeVito, Z.~Lin,
  A.~Desmaison, L.~Antiga, and A.~Lerer, ``Automatic differentiation in
  pytorch,'' 2017.

\bibitem{sgd}
L.~Bottou and O.~Bousquet, ``The tradeoffs of large scale learning,'' in {\em
  Advances in neural information processing systems}, pp.~161--168, 2008.

\bibitem{ren2018tieredimagenet}
M.~Ren, E.~Triantafillou, S.~Ravi, J.~Snell, K.~Swersky, J.~B. Tenenbaum,
  H.~Larochelle, and R.~S. Zemel, ``Meta-learning for semi-supervised few-shot
  classification,'' in {\em 6th International Conference on Learning
  Representations, {ICLR} 2018, Vancouver, BC, Canada, April 30 - May 3, 2018,
  Conference Track Proceedings}, OpenReview.net, 2018.

\bibitem{russakovsky2015imagenet}
O.~Russakovsky, J.~Deng, H.~Su, J.~Krause, S.~Satheesh, S.~Ma, Z.~Huang,
  A.~Karpathy, A.~Khosla, M.~Bernstein, {\em et~al.}, ``Imagenet large scale
  visual recognition challenge,'' {\em International journal of computer
  vision}, vol.~115, no.~3, pp.~211--252, 2015.

\bibitem{oreshkin2018tadam}
B.~Oreshkin, P.~R. L{\'o}pez, and A.~Lacoste, ``Tadam: Task dependent adaptive
  metric for improved few-shot learning,'' in {\em Advances in Neural
  Information Processing Systems}, pp.~721--731, 2018.

\bibitem{krizhevsky2009cifar}
A.~Krizhevsky, V.~Nair, and G.~Hinton, ``Cifar-10 and cifar-100 datasets,''
  {\em URl: https://www. cs. toronto. edu/kriz/cifar. html}, vol.~6, 2009.

\bibitem{leo}
A.~A. Rusu, D.~Rao, J.~Sygnowski, O.~Vinyals, R.~Pascanu, S.~Osindero, and
  R.~Hadsell, ``Meta-learning with latent embedding optimization,'' in {\em 7th
  International Conference on Learning Representations, {ICLR} 2019, New
  Orleans, LA, USA, May 6-9, 2019}, OpenReview.net, 2019.

\end{thebibliography}
\bibliographystyle{ieeetr}



\end{document}